\documentclass[a4paper,twoside]{article}

\usepackage{epsfig}
\usepackage{subcaption}
\usepackage{calc}
\usepackage{amssymb}
\usepackage{amstext}
\usepackage{amsmath}
\usepackage{amsthm}
\usepackage{multicol}
\usepackage{pslatex}
\usepackage{apalike}
\usepackage{algorithm2e}
\usepackage[bottom]{footmisc}

\usepackage{xspace}
\usepackage{tabularx}
\usepackage{booktabs}
\usepackage{hyperref}
\usepackage{lipsum}

\let\svthefootnote\thefootnote
\newcommand\blankfootnote[1]{%
  \let\thefootnote\relax\footnotetext{#1}%
  \let\thefootnote\svthefootnote%
}
\let\svfootnote\footnote
\renewcommand\footnote[2][?]{%
  \if\relax#1\relax%
    \blankfootnote{#2}%
  \else%
    \if?#1\svfootnote{#2}\else\svfootnote[#1]{#2}\fi%
  \fi
}

\newcommand{\memseg}{MemSeg\xspace}
\newcommand{\oursmixed}{\textit{In}\&\textit{Out}\xspace}
\newcommand{\ours}{DDPM\xspace}
\newcommand{\ksdd}{KSDD2\xspace}

\newcommand{\lora}{LoRA\xspace}

\newcommand{\AP}{AP\xspace}
\newcommand{\PRECISION}{Precision\xspace}
\newcommand{\RECALL}{Recall\xspace}
\newcommand{\sane}{normal\xspace}
\newcommand{\anomalous}{anomalous\xspace}
\newcommand{\resnet}{ResNet-50\xspace}

\newcommand{\segdec}{MixedSegdec\xspace}

\usepackage{SCITEPRESS}     

\begin{document}

\title{Diffusion-based Image Generation for In-distribution Data Augmentation in Surface Defect Detection}

\author{
\authorname{
Luigi Capogrosso\sup{*},
Federico Girella\sup{*},
Francesco Taioli\sup{*},
Michele Dalla Chiara,
Muhammad Aqeel,
Franco Fummi,
Francesco Setti,
and Marco Cristani
}
\affiliation{Department of Engineering for Innovation Medicine, University of Verona, Italy}
\email{name.surname@univr.it}
}

\keywords{Diffusion Models, Data Augmentation, Surface Defect Detection.}

\abstract{
In this study, we show that diffusion models can be used in industrial scenarios to improve the data augmentation procedure in the context of surface defect detection.
In general, defect detection classifiers are trained on ground-truth data formed by normal samples (negative data) and samples with defects (positive data), where the latter are consistently fewer than normal samples.
For these reasons, state-of-the-art data augmentation procedures add synthetic defect data by superimposing artifacts to normal samples.
This leads to out-of-distribution augmented data so that the classification system learns what is not a normal sample but does not know what a defect really is.
We show that diffusion models overcome this situation, providing more realistic in-distribution defects so that the model can learn the defect's genuine appearance.
We propose a novel approach for data augmentation that mixes out-of-distribution with in-distribution samples, which we call \oursmixed{}.
The approach can deal with two data augmentation setups: \textit{i)} when no defects are available (zero-shot data augmentation) and \textit{ii)} when defects are available,  which can be in a small number (few-shot) or a large one (full-shot).
We focus the experimental part on the most challenging benchmark in the state-of-the-art, i.e., the Kolektor Surface-Defect Dataset 2, defining the new state-of-the-art classification AP score under weak supervision of .782.
The code is available at \url{https://github.com/intelligolabs/in_and_out}.
}

\onecolumn \maketitle \normalsize \setcounter{footnote}{0} \vfill

\section{\uppercase{Introduction}} \label{sec:Intro}

\footnote[]{\sup{*} These authors contributed equally to this work.}

\begin{figure*}[t]
\centering
\includegraphics[width=.9\linewidth]{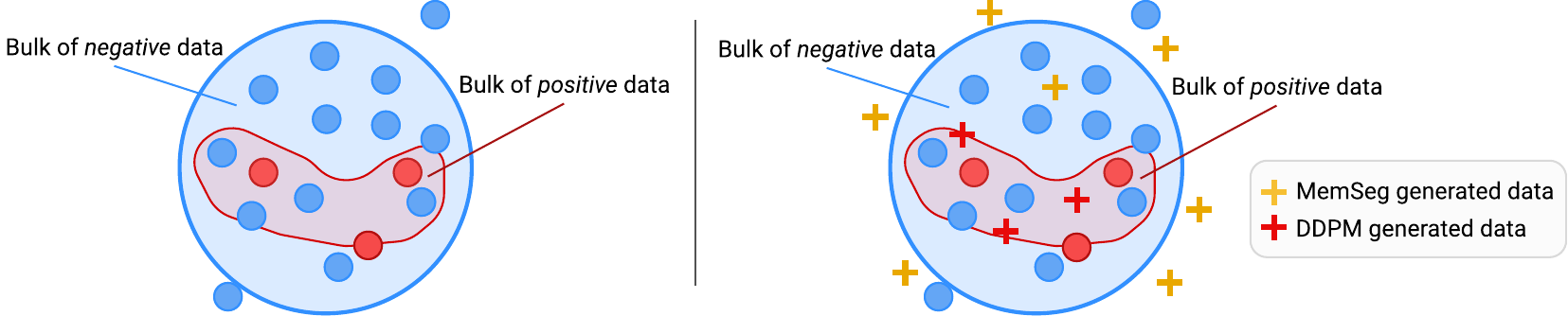}
\caption{Idea underlying our \oursmixed{} data augmentation approach.
(\emph{Left}, blue dots) The blue dots outside the bulk of negative data could be wrongly classified as anomalies (false positives), being slightly different from most of the negative data.
(\emph{Right}, yellow crosses) State-of-the-art per-region data augmentation methods (for example, \memseg{}~\cite{yang2023memseg}) add positive synthetic samples in that zone, which helps in deciding what is certainly not anomalous data.
(\emph{Left}, red dots) On the other hand, the red dot partially outside the bulk of positive data could be, in principle, understood as a negative sample, leading to a false negative.
(\emph{Right}, red crosses) Diffusion-based generated data is capable of producing defects very similar to the ones in the bulk of positive data, helping the classifier not produce false negative classifications.} 
\label{fig:teaser}
\end{figure*}

Surface defect detection is a challenging problem in industrial scenarios, defined as the task of individuating samples containing a defect~\cite{wang2018fast}.
The first solution involves hiring human experts: they check each product and remove the pieces with a defect.
Unfortunately, human experts can be biased and are subject to fatigue.
Instead, automated defect detection systems~\cite{tsang2016fabric,hanzaei2017automatic} solve the above problems by learning classifiers on defective and normal training samples. 
Unfortunately, data collection requires a strong human effort and extensive labeling times, and the collected data has a majority of normal samples (negative samples) since the defects (positive samples) are way less than the normal samples.
Training data becomes severely unbalanced in this general scenario, limiting the system's performance.

To solve this issue, data augmentation methodologies have emerged as viable solutions~\cite{zavrtanik2021draem,yang2023memseg,zhang2023prototypical}.
The main principle is to augment the real defective samples with synthetic ones to balance the class distributions.
To date, the best and most widely used approach for data augmentation consists of a per-region data augmentation~\cite{yang2023memseg}.
The idea is to start from real negative samples, overlaying them with regions containing texture artifacts, making them positive.
Unfortunately, this is far from being similar to a genuine defect, so we refer to that as out-of-distribution data.
In terms of detection precision, this approach works since these data are useful to indicate what is certainly not a normal sample, thus avoiding false positive classifications.
At the same time, this approach does little to avoid false negative classifications since defects are usually fine-grained deviations from normal samples, leading to low recall scores.

Diffusion models~\cite{dhariwal2021diffusion,rombach2022high} are deep generative models inspired by non-equilibrium thermodynamics that allow the sampling of rich latent spaces to generate meaningful realistic images.
In this paper, we promote using Denoising Diffusion Probabilistic Models (\ours{}s) to produce fine-grained realistic defects, solving the above issue.
Specifically, we can distinguish two different scenarios: \textit{i)} when no defects are available (zero-shot data augmentation); \textit{ii)} when some defects are available, which could be very few (few-shot, or $N$-shot with $N$ small) or in a large number (full-shot or $N$-shot with $N$ large).

In the first case, a human-in-the-loop paradigm is employed. 
Specifically, a human operator can drive the generation of proper defects by exploiting their domain knowledge.
This occurs using textual strings, which condition the generation of positive samples asking for specific defects (e.g., ``scratches'', ``holes'').
Instead, in the second scenario, when anomalous samples are available, fine-tuning can be done directly on them.
In this case, human operators are unnecessary since the model can already learn what a defect looks like.
In all the cases, we can observe that \ours{}-generated data is complementary to per-region augmented out-of-distribution data, as described in Figure~\ref{fig:teaser}, since it allows the enrichment of the statistics of positive data (in-distribution) ameliorating the downstream classification performance in terms of recall.

Due to the high complementarity of the two augmentation policies, we decided to use them together, dubbing our approach \oursmixed{} data augmentation since it is a compromise between augmented images that are in and out-of-distribution.
We test our approach on the Kolektor Surface-Defect Dataset 2 (\ksdd{})~\cite{bovzivc2021mixed}.
Notably, with 120 augmented images, the Average Precision (AP) classification score is .782, setting the new state-of-the-art performance on this dataset.

\section{\uppercase{Related Work}} \label{sec:SOTA}

One of the most adopted frameworks for automated quality control is defect detection, where the goal is to find images that contain defects.
Specifically, we focus on weakly supervised approaches~\cite{bovzivc2021mixed,zhang2021cadn}, in which positive and negative training images are labeled at the image level, that is, without per pixel masks.
This is the cheapest and most widely used annotation in industrial contexts.
 
Despite its importance and wide usage, the practice of data augmentation for defect detection received little attention in the literature, and this paper is one of the first that entirely focuses on it.

The most adopted pipeline for the generation of the anomalous synthetic samples consists of a series of random standard augmentations on the input image, such as mirror symmetry, rotation, brightness, saturation, and hue changes, followed by a super-imposition of noisy patches on the image~\cite{yang2023memseg,zhang2023diffusionad}.
Interestingly, in~\cite{zavrtanik2021draem}, an ablation study focused on the generation of synthetic anomalies leads to the following findings: \textit{i)} adding synthetic noise images is never counterproductive, it just diminishes the effectiveness in percentage; \textit{ii)} few generated anomaly images (in the order of tens) are enough to increase the performance substantially; \textit{iii)} textural injection in the anomalies is important, or, equivalently, adding uniformly colored patches is not effective.

In all of these papers, it is evident that the synthetically generated images are just out-of-distribution patterns, which do not have to represent the target-domain anomalies faithfully.
We improved this setup, being the first to focus on genuine in-distribution defect data.
A little improvement has been made in~\cite{zhang2023prototypical}, in which the authors introduced the concept of ``extended anomalies'', where the specific anomalous regions of the seen anomalies are placed at any possible position within the normal sample after having applied random spatial transformations. Unfortunately, this requires segmenting the training data, which we want to avoid.

\section{\uppercase{Background}} \label{sec:preliminaries}

We organize this section into four different parts, each one providing an overview of a topic related to our work: \textit{i)} \ours{}s; \textit{ii)} Dreambooth fine-tuning; \textit{iii)} Low-Rank Adaptation (\lora{}), and \textit{iv)} per-region data augmentation.

\paragraph{Denoising Diffusion Probabilistic Models.} \label{sec:diff_models}
\ours{}s are probabilistic models inspired by the non-equilibrium statistical physics phenomenon of diffusion~\cite{sohl2015deep,ho2020denoising}.
In recent years, diffusion models have gradually become state-of-the-art in image synthesis, surpassing GANs in performance~\cite{dhariwal2021diffusion}.
One of the main advantages of such models is the ability to guide the sampling steps with additional input data with a technique called conditioning.
The most common form of conditioning is a text that describes what the expected image should look like~\cite{rombach2022high}.
However, recent developments have explored other forms of conditioning, such as images, segmentation maps, or logic formulas~\cite{capogrosso2023neuro}.

\paragraph{Dreambooth fine-tuning.} \label{sec:dreambooth}
Dreambooth~\cite{ruiz2023dreambooth} is a procedure for \ours{}s that allows fine-tuning the model with a small number $N$ of images.
During the fine-tuning steps, each of the $N$ images is associated with a prompt defining the identification token and the subject class.
At the same time, regularization images (images of the same class but without the subject identification token) are used to prevent the fine-tuning model from forgetting the subject class learned during the original (non-fine-tuning) training, thanks to a prior preservation loss.
This allows the DDPM to learn a new specialized concept, represented by the identification token, with fewer iterations and without overwriting its prior knowledge.

\paragraph{Low-Rank Adaptation (\lora{}).} \label{sec:lora}
In recent years, fine-tuning Large Language Models (LLMs) has become prohibitively expensive due to the huge number of parameters.
In~\cite{hu2021lora}, the authors introduced Low-Rank Adaptation (\lora{}), a model-agnostic method of fine-tuning models in an efficient way.
\lora{} has the following advantages: \textit{i)} many small \lora{} modules for different tasks can be built by a single pre-trained model; \textit{ii)} optimizes only the injected, much smaller low-rank matrices, lowering the hardware requirements barrier; \textit{iii)} the final model, obtained by merging the original pre-trained model and the low-rank matrices, has no additional inference latency.

\paragraph{Per-region data augmentation.}
\begin{figure}[t]
\centering
\includegraphics[width=.9\linewidth]{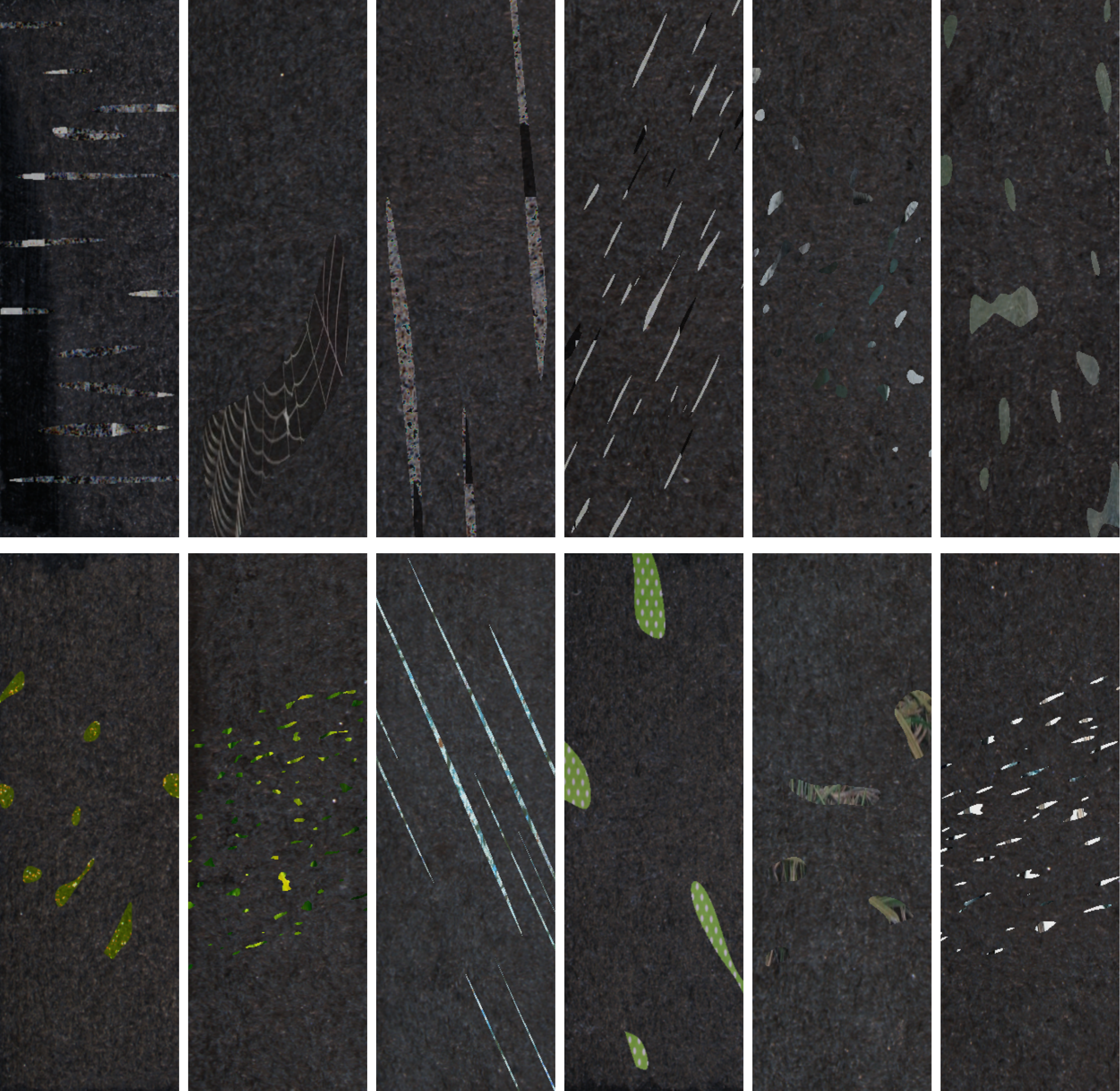}
\caption{Augmented images generated by the \memseg~\cite{yang2023memseg} pipeline.
It is evident how it provides out-of-distribution positive samples.}
\label{fig:imgs_memseg_daugs}
\end{figure}
With \textit{per-region} data augmentation, we refer to out-of-distribution data augmentation procedures that superimpose noise regions on the original image.
In our study, we will use \memseg{}~\cite{yang2023memseg} as our out-of-distribution data augmentation.
Some examples of images generated by the \memseg{} pipeline are reported in Figure~\ref{fig:imgs_memseg_daugs}.

\section{\uppercase{Method}} \label{sec:method}

\begin{figure*}[t]
\centering
\includegraphics[width=.95\linewidth]{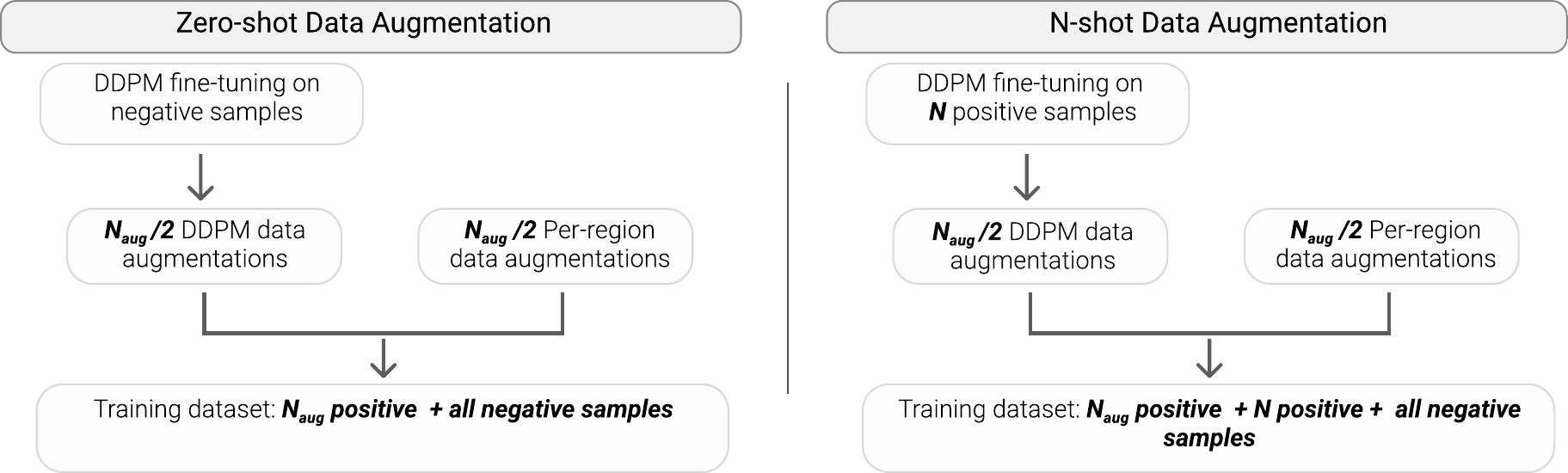}
\caption{General schema of our \oursmixed{} method.} 
\label{fig:schema}
\end{figure*}

The \oursmixed{} data augmentation aims at producing $N_{aug}$ additional positive images.
The approach can be applied, with slightly different pipelines, on two scenarios: \textit{i)} when no positive samples are available (zero-shot data augmentation) and \textit{ii)} when positive samples are available ($N$-shot data augmentation, where $N$ can be small or large).
In the following, the two pipelines are detailed; a graphical sketch is presented in Figure~\ref{fig:schema}.

\subsection{Zero-shot data augmentation} \label{sec:zeroshot_method}
In this scenario, we simulate that no positive samples are available in the training set. 
Thus, our aim is a zero-shot data augmentation procedure in which two steps are performed: fine-tuning and data augmentation.

\paragraph{Fine-tuning step.}
Dreambooth is adopted to perform fine-tuning on a \ours{}.
To reduce training time and lower computation requirements, we only train low-rank update matrices by employing \lora{}.
These update matrices are then summed to the original weights, completing the fine-tuning procedure.
Specifically, we control the weight of the \lora{} update matrices during the merge with a parameter $\alpha{}$: a value close to 0 results in no fine-tuning, while a value close to 1 results in the strongest fine-tuning.

In the zero-shot data augmentation, we perform fine-tuning with a portion of randomly chosen negative samples from the training set. 
The number of samples depends on the complexity of the data we want to manipulate: the larger the intra-class variance, the larger the number of elements to sample.
In this preliminary study, we select the number of samples heuristically (see Section~\ref{sec:exp} for details).

\paragraph{\textbf{Data augmentation.}}
In this step, we create the $N_{aug}$ augmented images generating $N_{aug}/2$ in-distribution images and $N_{aug}/2$ out-of-distribution images.
The $N_{aug}/2$ in-distribution images are obtained by exploiting the fine-tuned DDPM through natural language prompts, describing the desired anomalies.
To define the types of defects in natural language and verify how well text expressions are suited to generate a genuine defect for the data at hand, it is reasonable to perform some human-in-the-loop cycles, exploiting the expert's domain knowledge to evaluate the augmentation quality.
Specifically, the operator prompts textual expressions and evaluates the generated data (total of $N_{aug}/2$), certifying reasonable defects or revising expressions for improved generations.
The $N_{aug}/2$ out-of-distribution images are obtained by the per-region data augmentation, detailed in Section~\ref{sec:preliminaries}.

This ensures that half of the augmented data will be in-distribution, describing the visual appearance of the defects (the diffusion-based one), while the other half of the data will focus on specifying what is certainly not a perfect sample (the per-patch images).
After the augmentation, the final training dataset will be formed by $N_{aug}$ augmented positive images plus all the original negative samples.

\subsection{N-shot data augmentation} \label{sec:nshot}
In this scenario, we assume to have $N$ images from the positive pool of dataset images on which we perform Dreambooth fine-tuning with \lora{}.
We refer to the cases where $N\sim{}5$ as few-shot data augmentation.
After the fine-tuning, $N_{aug}/2$ in-distribution positive samples are generated. 
As for the zero-shot data augmentation scenario, the additional $N_{aug}/2$ out-of-distribution images are obtained by the per-region data augmentation, detailed in Section~\ref{sec:preliminaries}. 

After the augmentation, the final training dataset will be formed by $N_{aug}$ augmented positive images + $N$ original positive images plus the negative samples.

\section{\uppercase{Experiments}} \label{sec:exp}

In this study, we explore the efficacy of our \oursmixed{} data augmentation approach for defect detection on the \ksdd{} dataset.

\paragraph{Dataset.}
\begin{figure}[!t]
\centering
\includegraphics[width=\linewidth]{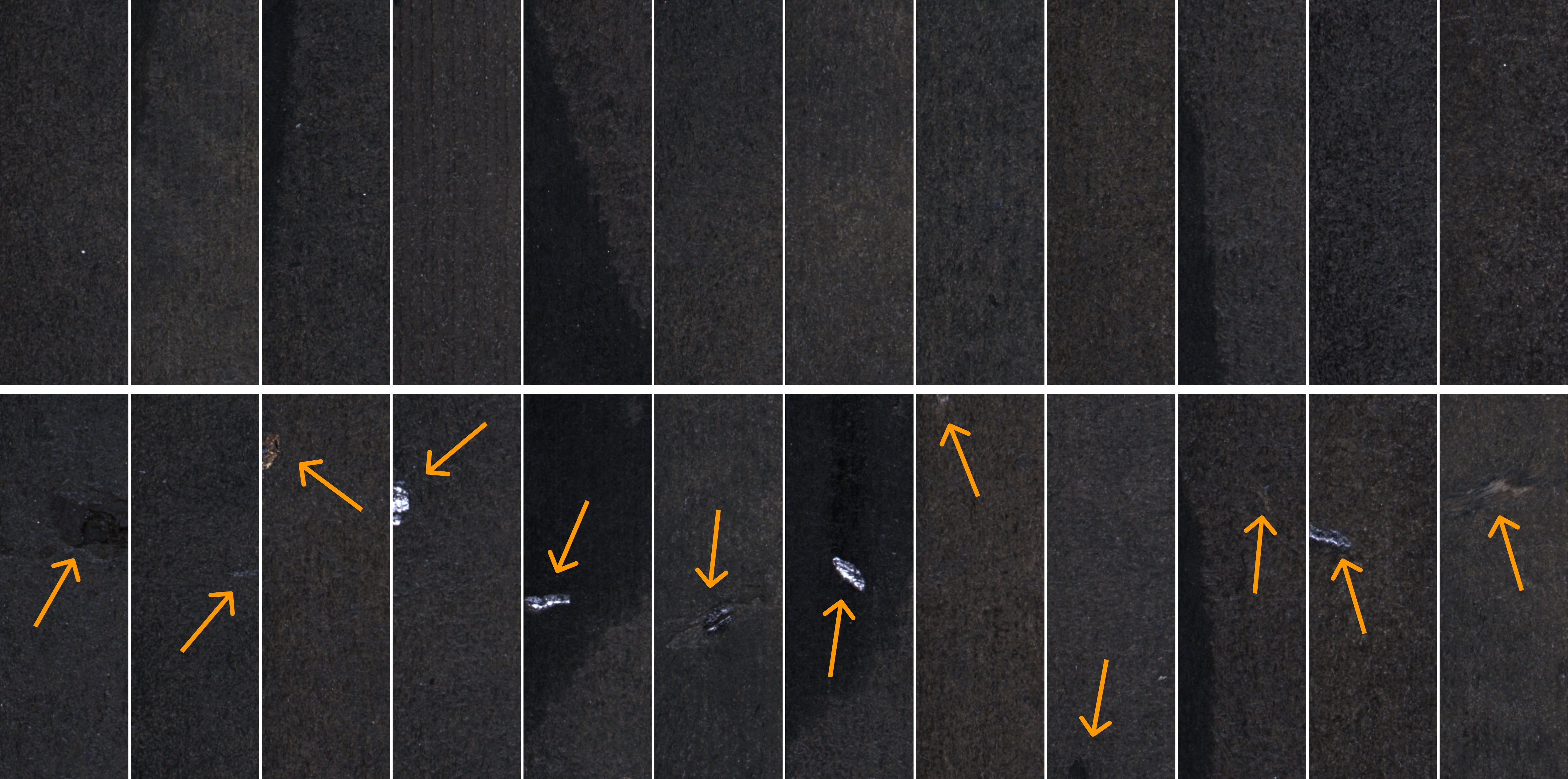}
\caption{Normal (top row) and anomalous (bottom row) samples from the \ksdd{} dataset. Note that some defects are very difficult to find.}
\label{fig:ksdd2_normal_anomaly}
\end{figure}

The \ksdd{} contains RGB images of defective production items, provided and annotated by Kolektor Group d.o.o.
The defects vary in shape, size, and color, ranging from small scratches and minor spots to large surface imperfections.

Since the images are of different sizes, we standardize the dataset resolution by center-cropping and resizing all the images to $200\times{}600$ pixels.
The dataset is split into train and test subsets, with 2085 negative and 246 positive samples in the training set, and 894 negative and 110 positive samples in the test set.
At the moment of writing, the state-of-the-art AP on this dataset stands at .733~\cite{bovzivc2021mixed}.
We show several \sane{} and anomalous samples in Figure~\ref{fig:ksdd2_normal_anomaly}.

\subsection{Implementation details} \label{sec:impl_details}

In this section, we specify all the implementation details for the sake of reproducibility.
All training and inferences have been carried out on an NVIDIA RTX 4090 GPU.

\paragraph{DDPM fine-tuning.}
In our experiments, we use Stable Diffusion~\cite{rombach2022high} as \ours{}.
The fine-tuning process follows the Dreambooth procedure (see Section~\ref{sec:dreambooth} for details).
We used the prompt ``\texttt{skt background}'', where ``\texttt{skt}'' is the identification token.
As written in Section~\ref{sec:preliminaries}, the string ``\texttt{skt}'' has no semantic meaning, and was selected to define an ID code for a new visual class. 
On the other hand, ``\texttt{background}'' is the subject class, identified as the most suited to obtain images with a homogeneous background.
The regularization images have been generated using the prompt ``\texttt{background}''.
The weight of the prior preservation loss is set to $1.0$ as in the original paper.
For faster training time and lower computation requirements, we also employ the \lora{}-c3Lier low-rank adaptation, a modified version of \lora{} that also applies low-rank approximations to $3\times{}3$ convolutional kernels and linear layers.

The code is implemented in PyTorch.
We used AdamW8bit~\cite{dettmers2022optimizers} as an optimizer, with a learning rate of $1e-5$.
We kindly direct the reader's attention to our configuration file for a more comprehensive exploration of the various hyperparameters involved.

\paragraph{DDPM data augmentation.}
\begin{figure}[!t]
\centering
\includegraphics[width=\linewidth]{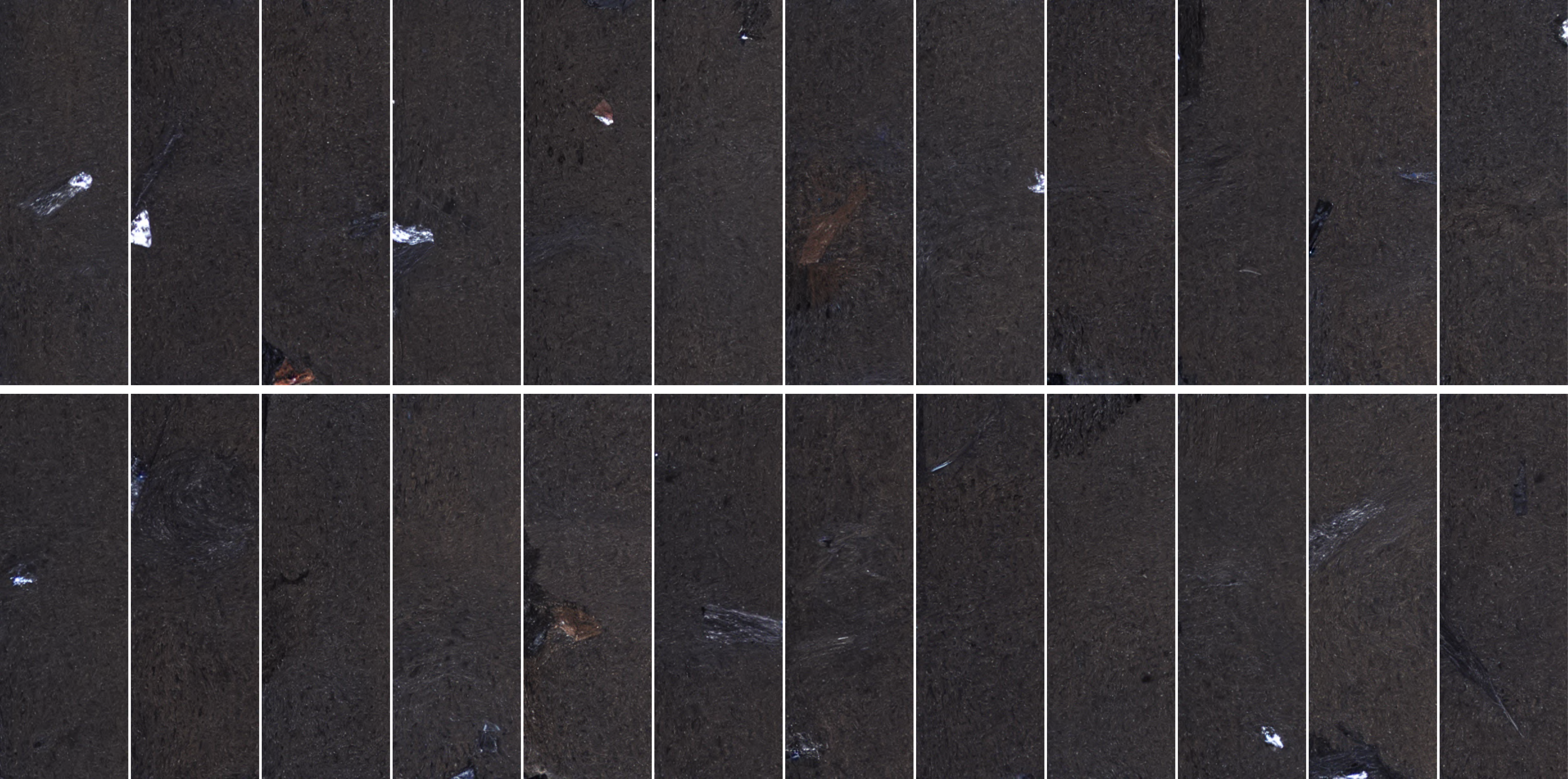}
\caption{Anomalous samples generated by \ours{}.
It is evident how it provides in-distribution positive samples.}
\label{fig:lora_zshot_c3lier_cropped}
\end{figure}
After training Stable Diffusion, we use it to generate $N_{aug}/2$ augmented images. 
In the zero-shot scenario, we use the prompts ``\texttt{skt background cracked}'' and ``\texttt{skt background scratched}'' to induce the generation of anomalous samples.
These prompts have been chosen after a series of tests and result in images containing plausible anomalies like the ones shown in Figure~\ref{fig:lora_zshot_c3lier_cropped}.
These generated images are then added to the training set, which will be used to train the anomaly detection model.
We train and evaluate this model with four different seeds for each of our experiments, generating $N_{aug}/2$ new images each time to provide the most statistically relevant results.

\paragraph{\resnet{} training and testing.}
We use the PyTorch implementation of the \resnet{}~\cite{he2016deep} as our anomaly detection model, in which we substitute the fully connected layers after the backbone to make it a binary classifier.
The network is trained for 50 epochs with an SGD optimizer, a learning rate of $0.01$, and a batch size of 5.

To keep consistency with the training and evaluation procedures of the \ksdd{}, we modify their official implementation to accommodate our \resnet{} model.
In particular, our setup is similar to the weakly supervised one presented in~\cite{bovzivc2021mixed}, where only the images and ground truth labels are used to train the model.
For each scenario, i.e., zero-shot data augmentation and \textit{N}-shot data augmentation, we will train three versions of our \resnet model: \textit{i)} using only \memseg{} to generate $N_{aug}$ images; \textit{ii)} using only our \ours{} to generate $N_{aug}$ images; and \textit{iii)} using \oursmixed{} as data augmentation, resulting in $N_{aug}/2$ images generated by \memseg{} and $N_{aug}/2$ generated by our \ours{}.

\subsection{Zero-shot data augmentation} \label{subsec:zero_shot}
\begin{table*}[ht]
\centering
\begin{footnotesize}
\caption{Results between \memseg{} and \ours{} when \textbf{\textit{no}} anomalous samples are available.}

\begin{tabular}{c|c|c|c||c|c|c|c}
\toprule
$\mathbf{N_{aug}}$ & \textbf{\AP{} $\uparrow$} & \textbf{\PRECISION{} $\uparrow$} & \textbf{\RECALL{} $\uparrow$} &
$\mathbf{N_{aug}}$ & \textbf{\AP{} $\uparrow$} & \textbf{\PRECISION{} $\uparrow$} & \textbf{\RECALL{} $\uparrow$} \\
\midrule
\memseg{} 80   & .514 (.026)  & \textbf{.733} (.113)  & .436 (.033)   & \ours{} 80   & \textbf{.547} (.086)  & .427 (.301)  & .695 (.194) \\ 
\memseg{} 100  & .388 (.066)  & .633 (.129)  & .432 (.054)            & \ours{} 100  & .532 (.028)  & .387 (.277)  & \textbf{.714} (.286) \\ 
\memseg{} 120  & .511 (.050)  & .683 (.054)  & .470 (.091)            & \ours{} 120  & .445 (.186)  & .465 (.329)  & .591 (.274)          \\
\midrule
Average  & .471 (.047)  & \textbf{.683} (.099)  & .446 (.059)      & Average  & \textbf{.508} (.100)  & .426 (.302)  & \textbf{.667} (.251)  \\
\bottomrule
\end{tabular}
\label{tab:lora_zero_shot}
\end{footnotesize}
\end{table*}
In these experiments, we emulate a situation where \textbf{\textit{no}} positive samples are available in the training set.
With this premise, we train our diffusion model with only $50$ randomly chosen negative samples from the training set.
We chose this number empirically and deemed it sufficient to represent the intra-class variance of the negative samples.
We train the \ours{} for 5 epochs, using as guiding prompt ``\texttt{skt background}'' and $\alpha{}=0.60$.

Once the diffusion model is trained, we generate $N_{aug}/2$ augmented positive samples using prompts specific to the dataset.
In our case, we used prompts such as ``\texttt{skt background cracked}'' and ``\texttt{skt~background~scratched}'', resulting in images like the ones shown in Figure~\ref{fig:lora_zshot_c3lier_cropped}. 
Therefore, we produce $N_{aug}/2$ out-of-distribution images by \memseg{}, obtaining the $N_{aug}$ of our \oursmixed{} approach.
We also experiment with fully-\memseg{} and fully-\ours{} augmentation pipelines for comparison.

We train the \resnet{} model on different values of $N_{aug}$ and evaluate it on the original test set. 
For each number of data augmentation, four different seeds have been used to report the most statistically relevant results.
We report the comparison between \memseg{} and \ours{} in Table~\ref{tab:lora_zero_shot}, where the numbers outside the parenthesis indicate the average results over the four seeds, while the numbers between parenthesis indicate the standard deviation.
As we can see, \ours{} achieves the highest AP (.547), recorded at 80 augmented images, while also resulting in an overall higher mean AP  when compared to the \memseg{} pipeline (.508 vs. .471).

\begin{figure}[!t]
\centering
\includegraphics[width=.90\linewidth]{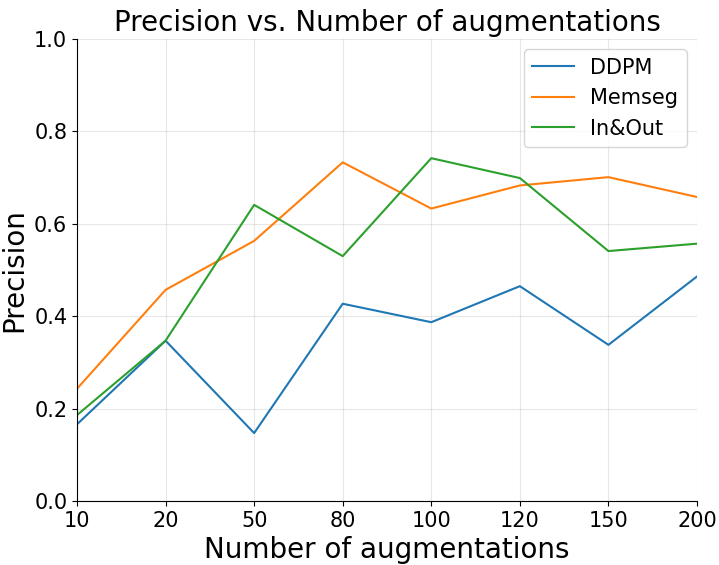}
\caption{Precision of the methods as a function of the number of augmentations.
Note that \memseg{} has higher overall precision. \oursmixed{} balances this metric.}
\label{fig:curve_1}
\end{figure}
\begin{figure}[!t]
\centering
\includegraphics[width=.90\linewidth]{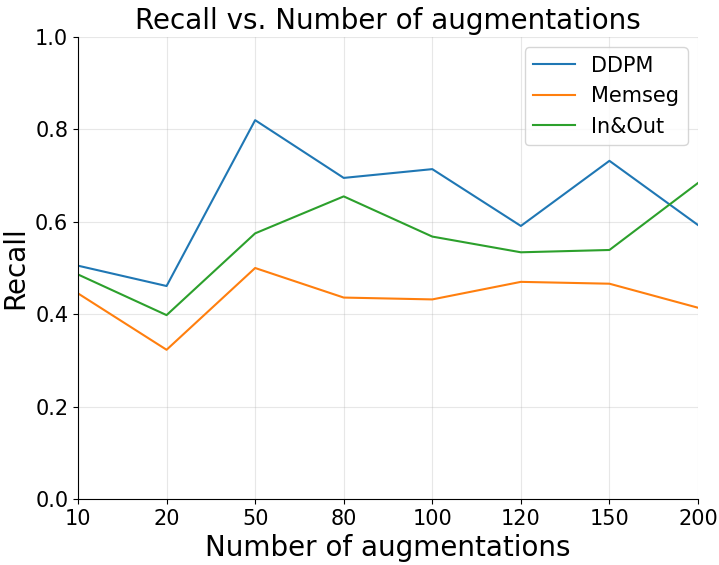}
\caption{Recall of the methods as a function of the number of augmentations.
 Note that \ours{} has a higher overall recall. \oursmixed{} balances this metric.}
\label{fig:curve_2}
\end{figure}
We want to highlight the difference between the precision and recall scores of \memseg{} and \ours{}.
While \ours{} achieves a higher recall (.714), the \memseg{} pipeline results in a higher precision (.733). 
This behavior is clearly shown in Figure~\ref{fig:curve_1} and~\ref{fig:curve_2}, where we plot the values of precision and recall of the two methods for different $N_{aug}$.

\begin{table}[ht!]
\centering
\begin{footnotesize}
\caption{Results when \textbf{\textit{no}} anomalous samples are available using \oursmixed{}. 
Thus, $N_{aug}/2$ samples generated with \ours{} and $N_{aug}/2$ with \memseg{}.}

\begin{tabular}{c|c|c|c}
\toprule
$\mathbf{N_{aug}}$ & \textbf{\AP{} $\uparrow$} & \textbf{\PRECISION{} $\uparrow$} & \textbf{\RECALL{} $\uparrow$} \\
\midrule
\oursmixed 80   & .556 (.085)                & .530 (.219)                 & \textbf{.655} (.065)  \\
\oursmixed 100  & \textbf{.626} (.059)       & \textbf{.742} (.109)        & .568 (.029)  \\
\oursmixed 120  & .536 (.023)                & .699 (.085)                 & .534 (.086)  \\
\midrule
Average & \textbf{.573} (.056)  & .657 (.138)  & .586 (.060) \\
\bottomrule
\end{tabular}
\label{tab:mixed_zero_shot}
\end{footnotesize}
\end{table}
When combined in the \oursmixed{} pipeline, where half of the augmented positive samples are provided by \ours{} and the other half is provided by \memseg{}, we obtain a huge performance boost in maximum (.626) and average (.573) \AP{}, with balanced precision and recall metrics.
These results, reported in Table~\ref{tab:mixed_zero_shot}, suggest how combining in-distribution (\ours{}) and out-of-distribution (\memseg{}) data, ameliorates precision and recall scores, helping the model better understand what an \anomalous{} sample is.

\subsection{$N$-shot data augmentation, $N$ small} \label{sec:few_shot}
Within manufacturing environments, organizations strive to minimize the occurrence of defects, resulting in a generally restricted number of anomalous samples. 
In this sub-section, we put ourselves in this situation, i.e., only a minimal amount of ground truth positive samples are available in the dataset.

\begin{table*}[ht!]
\centering
\begin{footnotesize}
\caption{Results between \memseg{} and \ours{} when \textbf{\textit{few}} anomalous images are available. Each training set contains $N=5$ anomalous samples, plus $N_{aug}$ augmented images.}

\begin{tabular}{c|c|c|c||c|c|c|c}
\toprule
$\mathbf{N_{aug}}$ & \textbf{\AP{} $\uparrow$} & \textbf{\PRECISION{} $\uparrow$} & \textbf{\RECALL{} $\uparrow$} &
$\mathbf{N_{aug}}$ & \textbf{\AP{} $\uparrow$} & \textbf{\PRECISION{} $\uparrow$} & \textbf{\RECALL{} $\uparrow$} \\
\midrule
\memseg 80{}   & .582 (.018)              & \textbf{.836} (.101)   & .466 (.049)  & \ours 80     & .580 (.045)     &  .542 (.270) & \textbf{.634} (.212)  \\
\memseg 100{}  & .511 (.086)               & .686 (.082)            & .527 (.069)  & \ours 100    & .526 (.075)              &  .610 (.063) & .477 (.081) \\ 
\memseg 120{}  & \textbf{.593} (.044)      & .801 (.065)            & .507 (.053)  & \ours 120    & .535 (.063)              &  .659 (.127) & .491 (.046) \\
\midrule
Average  & \textbf{.562} (.049)  & \textbf{.774} (.083)  & .500 (.057)    & Average     & .547 (.061)  & .604 (.153)  & \textbf{.534} (.113)  \\
\bottomrule
\end{tabular}
\label{tab:few_shot}
\end{footnotesize}
\end{table*}
To simulate this challenging setup, we randomly select only $N=5$ anomalous samples from the \ksdd{} training dataset and use them to fine-tune the \ours{} for 49 epochs with $\alpha{}=0.95$.
Following the procedure introduced in Section~\ref{sec:nshot}, we generate several training sets induced by the different $N_{aug}$ of new samples, plus the $N$ images on which we trained the \ours{}.
For the classifier, we use the same \resnet{} architecture. 
The findings of this experiment are documented in Table~\ref{tab:few_shot}.
As we can see, the \memseg{} method slightly outperforms \ours{}, resulting in an average AP of .562 and .547, respectively.
Moreover, \memseg{} produces a maximum AP of .593 at $N_{aug}=120$, while \ours{} records a maximum AP of .580 at $N_{aug}=80$.
The precision and recall have similar behavior as seen in~\ref{subsec:zero_shot}, with \ours{} having a higher recall (.634 vs. .527) and lower precision (.659 vs .836) w.r.t. \memseg{}.

\begin{table}[t!]
\centering
\begin{footnotesize}
\caption{Results when \textbf{\textit{few}} anomalous images are available using \oursmixed{}.
Each training set contains $N_{pos}=5$ anomalous samples, plus $N_{aug}$ augmented images, where half samples are generated by \ours{} and half by \memseg{}.}

\begin{tabular}{c|c|c|c}
\toprule
$\mathbf{N_{aug}}$ & \textbf{\AP{} $\uparrow$} & \textbf{\PRECISION{} $\uparrow$} & \textbf{\RECALL{} $\uparrow$} \\
\midrule
\oursmixed 80   & .531 (.041)                & .507 (.220)               & .655 (.126)  \\
\oursmixed 100  & \textbf{.578} (.041)       & .450 (.343)               & \textbf{.761} (.245)  \\
\oursmixed 120  & .575 (.025)                & \textbf{.635} (.316)      & .636 (.189)  \\
\midrule
Average & .561 (.036) & .531 (.293) & \textbf{.684} (.187) \\
\bottomrule
\end{tabular}
\label{tab:mixed_few_shot}
\end{footnotesize}
\end{table}
Interestingly enough, in Table~\ref{tab:mixed_few_shot}, we can see that the \oursmixed{} pipeline does not seem to increase the performance, achieving an average AP on par with \memseg{} (.561) while recording a slightly lower maximum AP (.578 vs. .593).
We hypothesize that, in this setup, \ours{} overfits the minimal number of anomalous images and cannot generalize the anomalous samples properly.
This is a problem if the samples on which we fine-tune the model are a subset of all the anomalies and, thus, are not representative enough of the entire anomalous distribution.

\subsection{$N$-shot data augmentation, $N$ large} \label{sec:full_shot}
\begin{table}[t!]
\centering
\begin{footnotesize}
\caption{Results when \textbf{\textit{all}} the anomalous samples are available using \oursmixed{}. 
Each training set contains all the anomalous \ksdd{} samples, plus $N_{aug}$ augmented images, where half of the samples are generated by \ours{} and half by \memseg{}.
Additionally, \oursmixed{} 0 indicates the performance achieved without data augmentation.
Note that \segdec{}~\cite{bovzivc2021mixed} indicates the results reported under the weakly supervised setting.}

\begin{tabular}{c|c|c|c}
\toprule
$\mathbf{N_{aug}}$ & \textbf{\AP{} $\uparrow$} & \textbf{\PRECISION{} $\uparrow$} & \textbf{\RECALL{} $\uparrow$} \\
\midrule
MixSegdec  & .733 (-)  & - (-)  & - (-)  \\
\oursmixed{} 0                     & .747 (.055)          & .826 (.081) & .723 (.058)  \\
\midrule
\oursmixed 80       & .747 (.022)    & .764 (.046)             & \textbf{.734} (.032)  \\
\oursmixed 100      & .775 (.013)    & .868 (.050)             & .720 (.026)  \\
\oursmixed 120      & \textbf{.782} (.030)   & \textbf{.906} (.064)    & .689 (.030)  \\
\midrule
Average & \textbf{.768} (.022)  & \textbf{.846} (.053)  & .714 (.029) \\
\bottomrule
\end{tabular}

\label{tab:mixed_full_shot}
\end{footnotesize}
\end{table}
Finally, to showcase \oursmixed{} as a general data augmentation technique, we explore the scenario with more positive samples in the training set. 
To this aim, we make all 246 positive samples available to the anomaly detection model during training, in addition to the usual $N_{aug}$ augmented anomalous images.
Following the procedure in Section~\ref{sec:nshot}, we use all the $N=246$ positive samples from the training set to fine-tune our diffusion model for 25 epochs with $\alpha{}=0.80$. 
Finally, we define a baseline by training the \resnet{} with $N_{aug}=0$ (\oursmixed{} 0), achieving an average AP of .747.
The results are reported in Table~\ref{tab:mixed_full_shot}. 

\begin{table*}[t!]
\centering
\begin{footnotesize}
\caption{Results between \memseg{} and \ours{} when \textbf{\textit{all}} the anomalous samples are available.}

\begin{tabular}{c|c|c|c||c|c|c|c}
\toprule
$\mathbf{N_{aug}}$ & \textbf{\AP{} $\uparrow$} & \textbf{\PRECISION{} $\uparrow$} & \textbf{\RECALL{} $\uparrow$} &
$\mathbf{N_{aug}}$ & \textbf{\AP{} $\uparrow$} & \textbf{\PRECISION{} $\uparrow$} & \textbf{\RECALL{} $\uparrow$} \\
\midrule
\memseg{} 80  & .744 (.007)           & .851 (.055)   & .691 (.058)   & \ours{} 80  & .758 (.007)           & .808 (.056)             & \textbf{.768} (.043) \\
\memseg{} 100 & \textbf{.774} (.016)  & .814 (.038)   & .752 (.028)   & \ours{} 100  & .763 (.008)          & .829 (.059)             & .725 (.034)  \\
\memseg{} 120 & .734 (.032)           & .772 (.107)   & .707 (.031)   & \ours{} 120  & .772 (.034)  & \textbf{.858} (.084)   & .725 (.061) \\
\midrule
Average & .751 (.018) & .812 (.067)   & .717 (.039)       & Average  & \textbf{.764} (.016)    & \textbf{.832} (.066)   & \textbf{.739} (.046) \\
\bottomrule
\end{tabular}
\label{tab:full_shot}
\end{footnotesize}
\end{table*}
The results of the two separate data augmentation procedures are reported in Table~\ref{tab:full_shot}.
In this scenario, the anomaly detection model trained with \ours{} augmented images achieves a maximum AP of .772, outperforming both the baseline (.747) and resulting in a higher average AP than \memseg{} (.764 vs. .751).
As we can see in Table~\ref{tab:mixed_full_shot}, \oursmixed{} achieves the highest average AP yet (.768) while balancing the precision and recall metrics, confirming our intuition.
Notably, with 120 augmented images, the maximum AP classification score is .782, beating the previous .733~\cite{bovzivc2021mixed} and setting the new state-of-the-art.

\section{\uppercase{Conclusion}} \label{sec:conclusion}

In this work, we introduce \oursmixed{}, a data augmentation method that generates positive images using \ours{}s for in-distribution samples and per-region augmentation for out-of-distribution samples.
We focus the experimental part on the \ksdd{}, defining the new state-of-the-art classification AP score under weak supervision of .782.
These results encourage further study on additional datasets and exploring how textual prompts interact with \ours{}, especially when defects are very few and not limited to cracks and scratches.

\section*{\uppercase{Acknowledgements}}
This study was carried out within the PNRR research activities of the consortium iNEST (Interconnected North-Est Innovation Ecosystem) funded by the European Union Next-GenerationEU (Piano Nazionale di Ripresa e Resilienza (PNRR) – Missione 4 Componente 2, Investimento 1.5 – D.D. 1058  23/06/2022, ECS\_00000043). This manuscript reflects only the Authors’ views and opinions, neither the European Union nor the European Commission can be considered responsible for them.

\bibliographystyle{apalike}
{\small
\bibliography{example}
}

\end{document}